\pdfoutput=1

\documentclass[11pt]{article}

\usepackage{ACL2023}

\usepackage{times}
\usepackage{latexsym}

\usepackage[T1]{fontenc}

\usepackage[utf8]{inputenc}

\usepackage{microtype}

\usepackage{inconsolata}

\usepackage{amsmath}
\usepackage{amsfonts}
\usepackage{amssymb}
\usepackage{bbm}
\usepackage{amsthm}
\usepackage{bbding}

\usepackage{booktabs}
\usepackage{multirow}

\usepackage{paralist}
\usepackage{varwidth}

\usepackage{mdwlist}
\usepackage{graphicx}

\usepackage{color}

\usepackage{xspace}
\usepackage{float}

\usepackage[ruled,linesnumbered]{algorithm2e}
\usepackage{tabularx}
\usepackage{xspace}
\usepackage{makecell}

\newcommand{\method}{NPPrompt\xspace}
\newcommand{\hide}[1]{}
\newcommand{\fullres}[2]{$#1_{#2}$}

\newcommand{\rebuttal}{\textcolor{black}}
\newcommand{\mask}{\texttt{[MASK]}\xspace}

\title{Pre-trained Language Models Can be Fully Zero-Shot Learners}

\author{Xuandong Zhao$^{\dagger}$, ~~ Siqi Ouyang$^{\dagger}$, ~~ Zhiguo Yu$^\ddagger$,  ~~ Ming Wu$^\ddagger$, ~~  Lei Li$^\dagger$  \\
  $^\dagger$UC Santa Barbara ~~~~ $^\ddagger$Microsoft \\
  \texttt{\{xuandongzhao,siqiouyang,leili\}@cs.ucsb.edu} \\
  \texttt{\{zhiguo.yu,mingwu\}@microsoft.com} 
}
\begin{document}
\maketitle

\begin{abstract}
How can we extend a pre-trained model to many language understanding tasks, without labeled or additional unlabeled data?
Pre-trained language models (PLMs) have been effective for a wide range of NLP tasks. 
However, existing approaches either require fine-tuning on downstream labeled datasets or manually constructing proper prompts. 
In this paper, we propose \textbf{n}on\textbf{p}arametric \textbf{prompt}ing PLM (\method) for fully zero-shot language understanding. 
Unlike previous methods, \method uses only pre-trained language models and does not require any labeled data or additional raw corpus for further fine-tuning, nor does it rely on humans to construct a comprehensive set of prompt label words. 
We evaluate \method against previous major few-shot and zero-shot learning methods on diverse NLP tasks: text classification, text entailment, similar text retrieval, paraphrasing, and multiple-choice question answering. 
Experimental results demonstrate that our \method outperforms the previous best fully zero-shot method by big margins, with absolute gains of 12.8\% in accuracy on text classification and 15.6\% on the GLUE benchmark.
\rebuttal{Our source code is available at \url{https://github.com/XuandongZhao/NPPrompt}.}
\end{abstract}

\section{Introduction}
\label{sec:intro}

%

Natural language understanding (NLU) has been important in many applications such as intelligent dialog assistants, online search, and social media analysis. 
Recent advancement of NLU has been driven by emergent pre-trained language models (PLMs) including BERT \citep{Devlin2019BERTPO, Liu2019RoBERTaAR}, GPT~\citep{radford2018improving, radford2019language, Brown2020LanguageMA}, BART \citep{Lewis2020BARTDS}, and T5 \citep{JMLR:v21:20-074}. 
Prior studies show that PLMs obtain substantial knowledge during pre-training on raw text corpus~\citep{Petroni2019LanguageMA, Feldman2019CommonsenseKM}.
By fine-tuning on task-specific labeled data, PLMs exploit such knowledge and gain impressive accuracy on a wide range of NLP tasks, such as text classification \citep{Kowsari2019TextCA}, question answering \citep{Rajpurkar2016SQuAD1Q}, machine reading comprehension \citep{Campos2016MSMA}, etc. 

However, fine-tuning approaches are expensive. 
It requires labeled datasets, which are rarely available for many tasks. 
Significant computational efforts are needed to update PLMs' parameters for multiple tasks. 
In addition, fine-tuning results in one distinct model for each task to maintain.


How can we generalize a pre-trained model to many NLP tasks, without labeled or additional unlabeled data?
Existing few-shot and zero-shot approaches propose to construct prompts to elicit desired predictions from PLMs~\citep{Brown2020LanguageMA}. 
The main idea of prompting PLMs is to convert an input utterance to one with masked templates. 
For example, in text classification an input can be ``The Warriors won the NBA championship 2022'' and it is instead converted to 
``The Warriors won the NBA championship 2022. This topic is about \texttt{[MASK]}''. 
A PLM (e.g. BERT) takes the converted text and produces predictions for the masked token, along with the probability. 
Ideally, a PLM will generate a higher probability for the word ``sports" than ``politics" on the \texttt{[MASK]} token.

\hide{
Unlike fine-tuning, prompt is the new prediction paradigm that freeze all parameters of the PLMs and directly model the probability of output text. By inserting a customized text template, it converts a specific task to a mask language modeling problem. Taking an example of news classification task, we could concatenate a prompt ``A \texttt{[MASK]} news: " to a news sentence (e.g., "The Warriors won the NBA championship 2022."). It would be natural to expect the PLM to generate a higher probability for the word ``sports" than ``politics" on the “\texttt{[MASK]}” token. Originated from GPT-3 \citep{Brown2020LanguageMA} and LAMA \citep{Petroni2019LanguageMA}, a series of studies using prompts \citep{Schick2021ExploitingCF, Schick2021ItsNJ, Gao2021MakingPL} for model tuning bridge the gap between pre-training objectives and down-stream tasks, and demonstrate that such discrete or continuous prompts induce better performances for PLMs on few-shot and zero-shot tasks.
}

Although these prompting-based methods are effective, they require unlabeled data for training or huge human efforts to construct prompts and to choose designated tokens to represent class labels \citep{Schick2021ExploitingCF,Schick2021ItsNJ,Gao2021MakingPL}. 
In addition, these manually constructed \textit{verbalizers}, i.e. mapping from words (e.g. ``basketball'') to class labels (e.g. \textsc{Sports}), do not extend to new emerging categories after PLMs are deployed.

\hide{
And they cannot easily be applied to the scenarios where new classes could be emerged after the learning stage \citep{RomeraParedes2015AnES}. For instance, the number of topics on social media is growing rapidly, there is no easy way to prepare each topic representation given labeled data or human domain knowledge are unfeasible to obtain for new topics \citep{Lee2011TwitterTT}. Therefore, a scalable zero-shot learning approach is needed to address this limited label data problem. 
}

\hide{traditional classification models used to recognize new topics can only use general information .} 

In this paper, we investigate the fully zero-shot learning problem for NLU where only the target label names are available but not the extra raw text. 
We propose \textbf{n}on\textbf{p}arametric \textbf{prompt}ing PLM (\method), a novel method to generate predictions for semantic labels without any fine-tuning. 
\method uses PLM's own embeddings to automatically find relevant words to labels (e.g. ``basketball'' and ``NBA'' for \textsc{Sports}), therefore it does not need humans to construct verbalizers. 
Our key idea is to search for the top $k$ nearest neighbors to a label name in the embedding manifold and then generate and aggregate PLM's predicted logits from masked prompts. 
In the above case, both predicted values for ``basketball'' and ``NBA'' contribute to the final prediction for the \textsc{Sports} category.
In this way, \method can be easily generalized to any new categories as long as the category names are semantically meaningful. 

The contributions of this paper are as follows. 
a) We develop \method, a novel method for fully zero-shot learning with PLMs. 
b) We conduct extensive experiments on diverse language understanding tasks including text classification, text entailment, similar text retrieval, paraphrasing, and multiple-choice question answering. 
Experimental results show that \method outperforms the previous zero-shot methods by absolute 12.8\% in accuracy on text classification and 15.6\% on the GLUE benchmark. 
Surprisingly, \method is on a par with the best prior method that trained with manual verbalizers, an additional knowledge base, and extra unlabeled data.

\hide{
Unlike manual verbalizers \citet{Schick2021ExploitingCF, Schick2021ItsNJ}, we design an automatic way to generate label words. For instance, in the above example, the mapping {sports} $\rightarrow$ \textsc{Sports} means that only predicting the word ``sports" for the \texttt{[MASK]} token is correct during inference, regardless of other related words such as ``basketball" and ``NBA". Therefore, to improve the coverage and apply the prompting to strict zero-shot text classification, we propose \method. Specifically, \method starts with the label words. We generate the word embedding for the label words using PLM's initial word embedding layer and search the whole vocabulary to find the top-$k$ words which share the most similar embedding spaces. Then we normalized these similarity scores as the weight of these expanded top-$k$ words. After we generated the predicted logits from the \mask~token ranging over the whole vocabulary, we only select a subset of scores with the same index of each label name's expanded top-$k$ words. In the end, we simply apply a weighted summation with those pre-calculated normalized similarities to represent each class's probability.
}

\section{Related Work}
\label{sec:related}
\paragraph{Prompting}
The success of GPT-3 \citep{Brown2020LanguageMA} has attracted much attention to prompting engineering, a new way to leverage pre-trained language models.
\cite{Brown2020LanguageMA} concatenate a few input and output pairs and feed them to the large-scale GPT-3 language model, which is an intuitive in-context learning paradigm, allowing the model to generate answers for additional cases autoregressively. 
Recent works \citep{Schick2021ExploitingCF, Schick2021ItsNJ} show that small-scale pre-trained language models such as BERT \citep{Devlin2019BERTPO}, RoBERTa \citep{Liu2019RoBERTaAR} and ALBERT \citep{Lan2020ALBERTAL} can also achieve decent performance using prompt-tuning. 
Prompting has been applied to a large variety of tasks such as Text Classification \citep{Schick2021ExploitingCF}, Natural Language Understanding \citep{Xu2022Go}, Knowledge Probing \citep{Petroni2019LanguageMA}, and Relation Extraction \citep{Han2021PTRPT}. 
Typically, a piece of prompt contains a template and a verbalizer. 
The language model predicts a probability distribution over vocabulary given the template and the verbalizer transforms it into a prediction over class labels. 
In this work, we focus on designing the verbalizers automatically.
\paragraph{Verbalizer Design}
The verbalizer plays a crucial role in prompting as it connects model outputs and labels, significantly influencing performance.
\cite{Schick2021ExploitingCF} design human written verbalizers for prompting, however, they are highly biased towards personal vocabulary with inadequate coverage. 
Apart from manually designed verbalizers, some recent studies explore automatic verbalizer construction. 
Auto-L \citep{Gao2021MakingPL} uses re-ranking to find the label words set by fine-tuning the model on the candidates searched by RoBERTa; 
AutoPrompt \citep{shin2020autoprompt} applies gradient-based search to create both prompts and label words automatically with a few trigger examples. 
But these approaches need to update parameters with gradient descent, which turns out to be infeasible without access to the model weights (e.g., GPT-3). 
KPT \citep{Han2021PTRPT} incorporates external knowledge into the verbalizer in which the unlabeled dataset is needed to refine the label words and thus is not applicable to scenarios where only label names are known. 
In contrast, our approach \method directly finds, without any gradient update, relevant words to label names with PLM's initial word embedding only.
\paragraph{Zero-shot Text Classification}
General zero-shot text classification typically focuses on classifying texts into categories that were not seen during the training process.  Transferring knowledge from seen classes to unseen ones requires accurate and discriminative descriptions of all classes \citep{Liu2019ReconstructingCN, Xia2018ZeroshotUI} or joint embeddings of categories and documents \citep{Nam2016AllinTL}. However, these methods rely on supervised data for the known label set, making them unsuitable for scenarios where no labeled pairs for any category are available.  SimPTC \citep{Fei2022BeyondPM}  improves zero-shot classification by clustering input texts and employing class-related prompts. LOTClass \citep{Meng2020TextCU} proposes a model that utilizes label names with self-training for zero-shot classification. Nonetheless, both SimPTC and LOTClass still require an unlabeled corpus or knowledge base to extract topic-related words and perform self-training. In contrast, \method achieves comparable or even superior performance without the need for any unlabeled dataset or knowledge base.


\begin{figure*}[t!]
\centering
\includegraphics[width=0.85\textwidth]{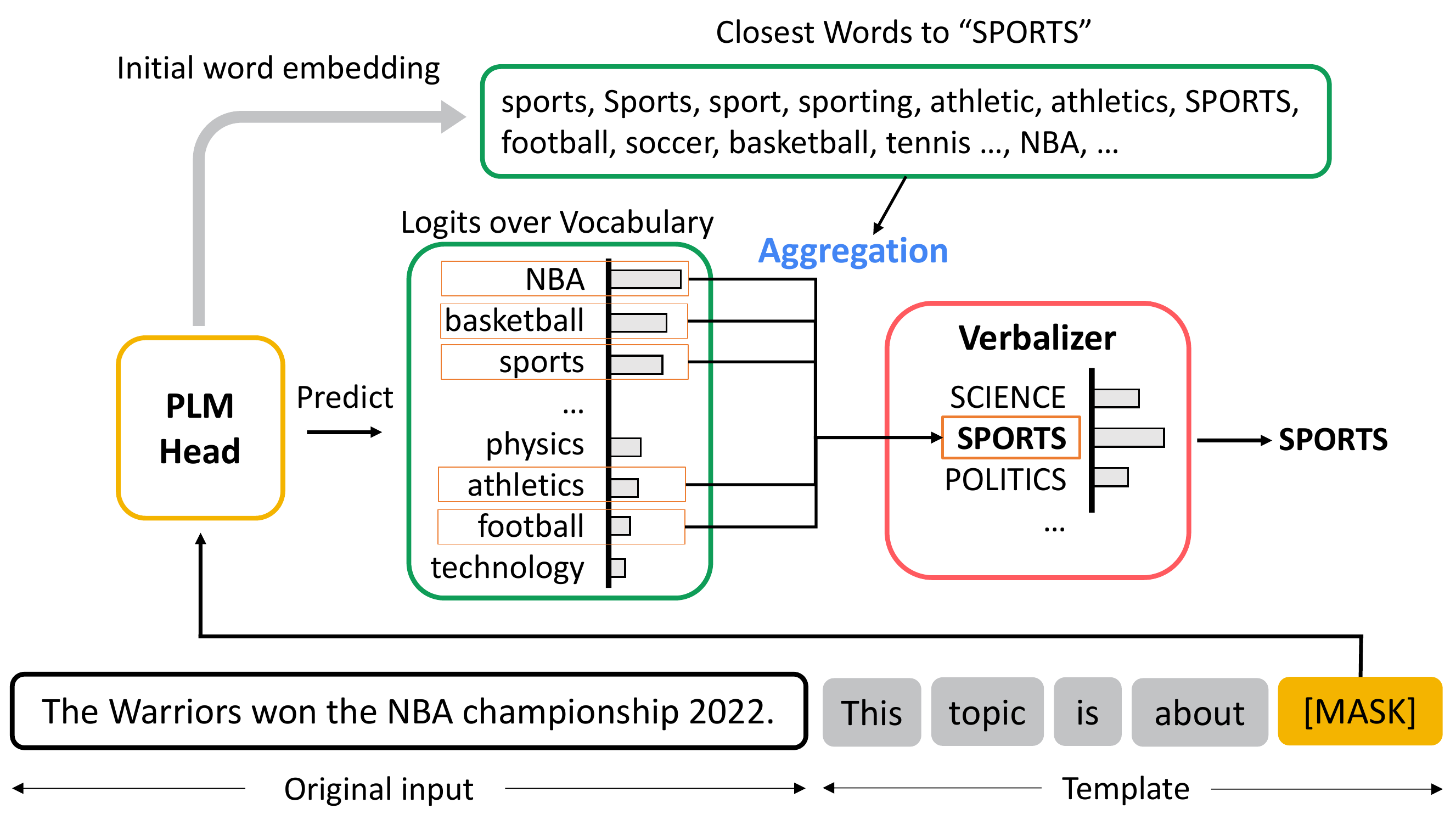}
\caption{The illustration of \method. We generate the label words by searching the related words from the initial word embedding of the pre-trained language model. By aggregating logits from the label words, we predict the category with the largest score (\textsc{Sports}).}
\label{fig:overview}
\end{figure*}

\section{Background: Prompt-based Tuning for PLMs}
\label{sec:background}

We first provide standard paradigms, prompt-based tuning, that perform well in few-shot scenarios, before introducing our approach for the zero-shot case. Take $N$ way text classification as an example. We aim to predict the label $y \in \mathcal{Y}$ for each sentence, where $\mathcal{Y}$ is the label set with $N$ distinct classes.

\hide{\subsection{Fine-tuning PLM}
With the advent of large-scale PLMs, fine-tuning approaches have been vital and successful in text classification tasks. 
Typically, for each text input $x = (t_1, t_2, \dots)$, we add a \texttt{[CLS]} token to the beginning and a \texttt{[SEP]} token to the end. Then the PLM encodes it into hidden representations $h = (h_{\texttt{[CLS]}}, h_1, h_2, \dots, h_{\texttt{[SEP]}})$. We can add extra classifier $F$ on top of the $h_{\texttt{[CLS]}}$ for various downstream tasks:
\begin{equation}
P(\cdot \mid x)=\operatorname{Softmax}\left(F\left(h_{\texttt{[CLS]}}\right)\right).
\end{equation}
The classifier and PLM are tuned by maximizing $\frac{1}{N} \sum_{i=1}^N \log P\left(y_i \mid x\right)$, where $y_i$ is the label of $x$.}

Prompt-based tuning tunes PLM using customized prompts \citep{Brown2020LanguageMA}. The regular prompt-based tuning converts a specific task to a cloze-style mask language modeling problem. For each input example $x$ (single sentence or sentence pair), we first apply a task template $\mathcal{T}$ on it, converting original input $x$ to $x_{\mathrm{prompt}}$. For instance, we concatenate the template ``$\mathcal{T}(\cdot) = \text{This topic is about } \mask$" with the original input ``The Warriors won the NBA championship 2022" and wrap it into:
\begin{equation*}
   x_{\mathrm{prompt}} = \mathcal{T}(x) = x \text{. This topic is about } \mask 
\end{equation*}
The \textit{verbalizer} $f$ in vanilla prompt engineering maps a set of selected words $\mathcal{V}$ from the vocabulary to the original label space $\mathcal{Y}$, i.e., $f: \mathcal{V} \rightarrow \mathcal{Y}$. 
Inversely, we use $\mathcal{M}(y_j)$ to denote the \textit{label words} in $\mathcal{V}$ that are mapped into a specific label $y_j$, $\cup_{y_j \in \mathcal{Y}} \mathcal{M}(y_j)=\mathcal{V}$. Then we calculate the probability of label $y_j$: 
\begin{small}\begin{equation*}
    P(y_j \mid x)=g\left(P(\texttt{[MASK]}=v_i \mid x_{\mathrm{prompt}} ) \mid v_i \in \mathcal{M}(y_j)\right),
\end{equation*}\end{small}
where $g(\cdot)$ is for aggregating the probability of label words into the probability of the label. Then PLMs can be fine-tuned by minimizing the cross-entropy loss with supervised examples.

\section{Proposed Method: \method}
\label{sec:approach}

We inherit PLM with verbalizers framework but keep PLM's parameters frozen \citep{Gao2021MakingPL}.
The key idea of \method is using PLM's word embeddings to automatically construct verbalizers -- mapping from words to labels -- in a fully zero-shot way.
It does not need any additional raw text corpus for fine-tuning. 
\method consists of two steps to compute predictions for any labels in a nonparametric form (Figure \ref{fig:overview}). 
1) We  search for all label words closely related to each class $y_j$ in PLM's token embedding manifold. 
2) Then we use the PLM to predict values for \mask, filter them using each class's set of label words, and aggregate the properly weighed outputs to produce the final prediction. 
In the following, we describe \method for text classification but it generalizes to other language understanding tasks. 

\paragraph{$k$-Nearest-Neighbor Verbalizer Construction}
For each class label (e.g. ``SPORTS''), we search over the whole vocabulary $\mathcal{V}$ for the top-$k$ words nearest to the label name in the PLM's embedding space. 
Here, the distance between words and label names is measured using  the cosine similarity score. 
Other distance metrics work as well and are examined in Section~\ref{sec:exp}.
We denote $k$ as the \textit{neighborhood number}. 
Assuming the embeddings of word $v_i$ and label name $y_j$ are $\mathbf{emb}(v_i)$ and $\mathbf{emb}(y_j)$ respectively, the label words of the verbalizer for $y_j$ are selected by top-$k$ ranking: 
\begin{equation}
\label{eq:topk}
\mathcal{M}(y_j) = \underset{v_i \in \mathcal{V}}{\operatorname{Top-}\!k}\left\{S(\mathbf{emb}(v_i), \mathbf{emb}(y_j))\right\},
\end{equation}
where $S(\cdot)$ is the cosine similarity function: $ S\left(\mathbf{emb}(v_i), \mathbf{emb}(y_j)\right)=\frac{\mathbf{emb}(v_i)}{\left\|\mathbf{emb}(v_i)\right\|} \cdot \frac{\mathbf{emb}(y_j)}{\left\|\mathbf{emb}(y_j)\right\|}$.

Since the PLM is already pre-trained on raw text corpus, it acquires sensible semantic knowledge and relatedness of words in the vocabulary. 
We use PLM's embedding to search for label words semantically relevant to given label names. 
For illustration, we show the found label words of two categories in the AG News dataset \citep{Zhang2015CharacterlevelCN} and the corresponding similarity scores in Table \ref{table:sim}. \rebuttal{We also extend our verbalizer to support label names with longer expressions in Appendix \ref{sec:multi-word}.}

\begin{table}[htbp]
\small
\centering
\begin{tabular}{llll}
\Xhline{2\arrayrulewidth} 
\multicolumn{1}{c}{\textbf{Word}} & \multicolumn{1}{c}{\textbf{Sim}} & \multicolumn{1}{c}{\textbf{Word}} & \multicolumn{1}{c}{\textbf{Sim}}  \\
\hline
`` sports" & 1.00 & `` business" & 1.00\\	  
`` Sports" & 0.77 & `` Business" & 0.78\\	
`` sport" & 0.75 & `` businesses" & 0.74\\	
`` sporting" & 0.68 & ``business" & 0.72\\	
`` athletics" & 0.65 & ``Business" & 0.67\\
``sports" & 0.65 & `` businessman" & 0.59\\
``Sports" & 0.65 & `` corporate" & 0.58\\
`` Sport" & 0.62 & `` company" & 0.56\\		
`` athletic" & 0.61	& `` enterprise" & 0.55\\
`` athletes" & 0.61 & `` businessmen" & 0.55\\
\Xhline{2\arrayrulewidth} 
\end{tabular}
\caption{The top 10 similar words of the RoBERTa-large model for the AG News dataset categories \textsc{Sports} and \textsc{Business}. Sim: cosine similarity scores.}
\label{table:sim}
\end{table}
\paragraph{Nonparametric Aggregation of Prompted Predictions}
For each input text $x$, we construct a prompt-augmented sequence $x_{\mathrm{prompt}}=\mathcal{T}(x)$ with a \mask token. 
We use the PLM to predict tokens for \mask.
In contrast to previous prompting methods which directly calculate the probability over the surface labels, 
we use the nearest label words from above to compute the probability for each output label. 
Only the words in a label's top-$k$ neighborhood will contribute to the class prediction. 
The contribution from each label word is non-equal. 

To be specific, with $\mathcal{T}(x)$, a PLM produces the logit vector $\Theta_{\texttt{[MASK]}}$ for all possible words at the \texttt{[MASK]} token. 
Notice that if the whole vocabulary is $\mathcal{V}$, $\Theta_{\texttt{[MASK]}} \in \mathbb{R}^{|\mathcal{V}|}$.
Then we compute the class probability for a label $y_j$ by aggregating the logits filtered by the verbalizer's label words. 
We use kernel smoothing to aggregate as follows:
\begin{equation}
\label{eq:logits}
\small
    Q(y_j|x)\!=\!\sum_{v_i\!\in\!\mathcal{M}(y_j)}\!w(v_i, y_j) \cdot \Theta(\texttt{[MASK]} \!=\!v_i| x_{\mathrm{prompt}}\!=\!\mathcal{T}(x))
\end{equation}
Where the weight between label word $v_i$ and class name $y_j$ is defined as:
\begin{equation}
\label{eq:weight}
\small
    w(v_i, y_j) = \frac{\operatorname{exp}\left(S(\mathbf{emb}(v_i), \mathbf{emb}(y_j))\right)}{\sum_{v_t \in \mathcal{M}(y_j)} \operatorname{exp} \left(S(\mathbf{emb}(v_t), \mathbf{emb}(y_j))\right)}
\end{equation}
Finally, the best class prediction is selected from the maximum of all labels:
\begin{equation*}
\widetilde{y}=\underset{y_j}{\arg\!\max}~Q\left(y_j \mid x\right) .
\end{equation*}
Notice since we use kernel smoothing on logits instead of probability, $Q$ is also unnormalized probability. 

For example, AG News has two classes
$y_1 = $ \{\textsc{Science}\}, $y_2=$  \{\textsc{Sports}\}. 
From Table~\ref{table:sim}, the verbalizer for \textsc{Sports} $\mathcal{M}(y_1)$ includes label words ``sports'', ``athletics'', etc, and the verbalizer for \textsc{Business} $\mathcal{M}(y_2)$ includes label words ``business'', ``corporate'', etc. 
Given an input text $x$ ``The Warriors won the NBA championship 2022'', the prompt-augmented sequence $x_{\mathrm{prompt}}$ will be ``The Warriors won the NBA championship 2022. This topic is about \mask''.
The PLM computes logits for every word $\Theta(\texttt{[MASK]} = v| x_{\mathrm{prompt}})$.
\method computes the unnormalized probabilities for \textsc{Sports} and \textsc{Business}:
\begin{figure}[h]
    \centering
    \includegraphics[width=0.47\textwidth]{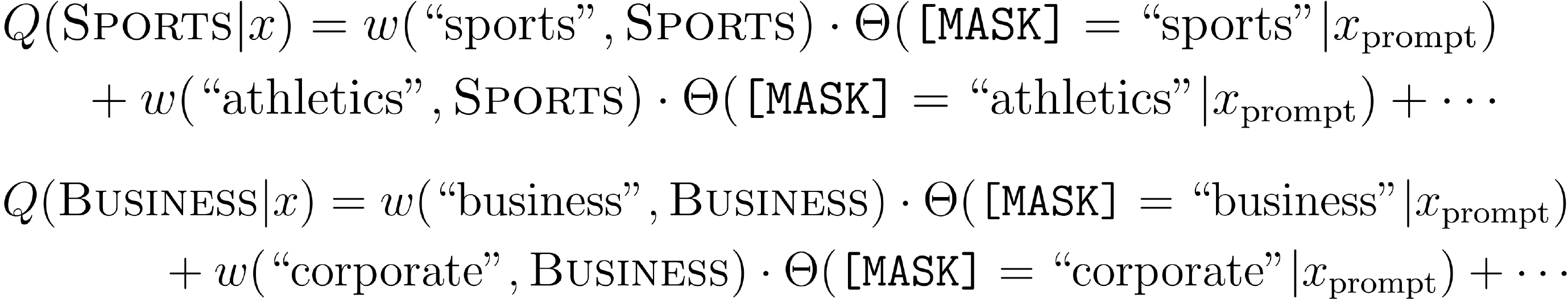}
\end{figure}

If the aggregated prediction $Q$ for \textsc{Sports} is larger than \textsc{Business}, \method outputs \textsc{Sports}. 

There are certain conditions where one class has label names containing little semantic meaning or where several keywords are needed to define a label. For instance, in the DBPedia dataset \citep{Lehmann2015DBpediaA}, one class is related to \textsc{NaturalPlace}, then we can use the keywords \{``river'', ``lake'', ``mountain''\} to represent this class. In this setting, we pick out the keyword with the maximum score calculated by Equation \ref{eq:logits} to represent each label first. Then we choose the label with the largest score. We use $\Phi(y_j)$ to denote all keywords in class $y_j$, and the final prediction is :
\begin{equation}
    \widetilde{y}=\underset{y_j}{\arg \max } \Big(\underset{y' \in \Phi(y_j)}{\arg \max }~Q\left(y' \mid x \right) \Big).
\label{eq:key_words}
\end{equation}

\section{Experiment}
\label{sec:exp}
We conduct extensive zero-shot learning experiments to demonstrate the effectiveness of our method. We provide detailed information on our implementation and address several research questions related to \method.

\subsection{Datasets, Prompt Templates, and Experimental Setup}
\begin{table}[h]
\setlength{\tabcolsep}{3pt} 
\small
\centering
\begin{tabular}{cccc} \\ 
\Xhline{2\arrayrulewidth}   
\textbf{Dataset} & \textbf{Classification Type} & \textbf{\# Classes} & \textbf{\# Test} \\
\hline
AG News & News Topic & 4 & 7,600 \\
DBPedia & Wikipedia Topic & 14 & 70,000 \\
IMDB & Movie Review Sentiment & 2 & 25,000 \\
Amazon & Product Review Sentiment & 2 & 400,000 \\
\Xhline{2\arrayrulewidth} 
\end{tabular}
\caption{Dataset statistics.}
\label{table:dataset}
\end{table}
We adopt sentiment classification tasks on two datasets, IMDB \citep{Maas2011LearningWV} and Amazon \citep{McAuley2013HiddenFA}, and topic classification tasks on another two datasets, AG News \citep{Zhang2015CharacterlevelCN} and DBPedia \citep{Lehmann2015DBpediaA}. All datasets are in the English language. For each task, we directly use the test set to assess model performances, without incorporating validation or training sets for post-tuning or cherry-picking hand-crafted prompts. The statistics of each dataset are shown in Table \ref{table:dataset}.

To concentrate on the verbalizer and reduce the influence of templates, we adopt multiple fixed manual templates following \cite{Hu2022KnowledgeablePI}. We report the best template used for the RoBERTa-large model in Table \ref{tab:template}.

\begin{table}[h]
\small
\setlength{\tabcolsep}{3pt} 
\centering
\begin{tabular}{cc} \\ 
\Xhline{2\arrayrulewidth}   
\textbf{Dataset} & \textbf{Template}  \\ 
\hline
AG News & A \texttt{[MASK]} news : $x$ . \\
DBPedia & $x_1$ $x_2$ In this sentence, $x_1$ is a \texttt{[MASK]} . \\
IMDB & $x$ All in all, it was \texttt{[MASK]} . \\
Amazon & $x$ All in all, it was \texttt{[MASK]} .  \\
\Xhline{2\arrayrulewidth} 
\end{tabular}
\caption{Prompt templates for \method.}
\label{tab:template}
\end{table}

We implement our experiments based on an open-source toolkit OpenPrompt \citep{ding2021openprompt}, which aims to conduct prompt learning easily. We choose RoBERTa-large \citep{Liu2019RoBERTaAR} as our pre-trained language model. We report the best accuracy of classification results for all experiments using different neighborhood numbers. Since we directly use the pre-trained models for testing, there is no randomness (random seed) in this process. All experiments are conducted on Nvidia A6000 GPUs and more details can be found in Appendix \ref{sec:app_exp}.

\begin{table*}[h]
\setlength{\tabcolsep}{3pt} 
\centering
\resizebox{0.85\textwidth}{!}{
\begin{tabular}{lccccccc}
\Xhline{2\arrayrulewidth} 
\textbf{Method}  & \textbf{Human/KB} & \textbf{Unlabeled} & \textbf{AG News} & \textbf{DBPedia} & \textbf{IMDB} & \textbf{Amazon} & \textbf{Avg.} \\
\hline
ManualVerb       & \CheckmarkBold & \XSolidBrush        & \fullres{79.6}{0.6} & \fullres{71.7}{1.1} & \fullres{92.0}{0.7} & \fullres{87.3}{0.4} & 82.7  \\
Semantic Retrieval   & \CheckmarkBold & \XSolidBrush    & \fullres{73.1}{1.2} & \fullres{78.6}{0.8} & \fullres{64.8}{1.3} & \fullres{59.4}{0.7} & 69.0  \\
NSP-BERT         & \CheckmarkBold & \XSolidBrush        & \fullres{77.4}{0.6} & \fullres{64.7}{5.3} & \fullres{72.8}{1.1} & \fullres{72.7}{3.9} & 71.9  \\
GPT-3 w. descriptions   & \CheckmarkBold & \XSolidBrush & 83.4 & 82.5 & 88.8 & 89.4 & 86.0  \\
ChatGPT w. descriptions & \CheckmarkBold & \XSolidBrush & 83.8 & 92.0 & 92.7 & \textbf{95.8} & 91.1 \\
SimPTC  & \CheckmarkBold & \XSolidBrush & \fullres{\textbf{86.9}}{0.3} & \fullres{\textbf{93.2}}{1.0} & \fullres{91.0}{0.0} & \fullres{93.9}{0.0} & \textbf{91.3} \\
LOTClass w/o. self train & \XSolidBrush & \CheckmarkBold & 82.2 & 86.0 & 80.2 & 85.3 & 83.4  \\
LOTClass        & \XSolidBrush & \CheckmarkBold         & 86.4 & 91.1 & 86.5 & 91.6 & 88.9  \\
KPT         & \CheckmarkBold & \CheckmarkBold             & 86.7 & 87.4 & \textbf{94.0} & 94.6 & 90.7  \\
\hline 
Null Prompt      & \XSolidBrush & \XSolidBrush        & \fullres{67.9}{2.0} & \fullres{56.8}{3.9} & \fullres{82.5}{1.5} & \fullres{89.4}{1.0} & 74.2  \\ 
Multi-Null Prompt  & \XSolidBrush & \XSolidBrush      & \fullres{68.2}{1.8} & \fullres{67.6}{1.8} & \fullres{86.6}{0.6} & \fullres{86.2}{2.7} & 77.2  \\
\method   & \XSolidBrush & \XSolidBrush & \fullres{\textbf{85.2}}{0.5} & \fullres{\textbf{86.8}}{0.1} & \fullres{\textbf{94.2}}{0.2} & \fullres{\textbf{93.9}}{0.0} & \textbf{90.0}  \\
\Xhline{2\arrayrulewidth} 
\end{tabular}
}
\caption{\rebuttal{Classification performance on four datasets with average results and standard error.} Human: with human efforts to write deceptions or design label words. KB: with external knowledge base; Unlabeled: with unlabeled corpus. Notice that our method achieves the best performance in a fully zero-shot setting, with an absolute improvement of 12.8\%. Surprisingly, it even approaches the best result with human effort/knowledge base and extra raw data.}
\label{table:main_res}
\end{table*}

\begin{table*}[h]
\setlength{\tabcolsep}{3pt} 
\centering
\resizebox{0.85\textwidth}{!}{
\begin{tabular}{lccccccccc}
\Xhline{2\arrayrulewidth} 
& \textbf{MNLI}  & \textbf{MNLI-mm} & \textbf{SST-2} & \textbf{QNLI}  & \textbf{RTE}  & \textbf{MRPC} & \textbf{QQP} & \textbf{CoLA} & \multirow{2}*{\textbf{Avg.}} \\
       & (acc) & (acc)   & (acc) & (acc) & (acc)& (F1) & (F1) & (Matt.) & \\ 
\hline
\multicolumn{10}{l}{\textit{With human designed prompts / few-shot data}}\\
Manual Label & 50.8 & 51.7 & 83.6 & 50.8 & 51.3 & 61.9 & 49.7 & 2.0 & 50.2 \\
In-context learning & \fullres{\textbf{52.0}}{0.7} & \fullres{\textbf{53.4}}{0.6} & \fullres{84.8}{1.3} & \fullres{53.8}{0.4} & \fullres{60.4}{1.4} & \fullres{45.7}{6.0} & \fullres{36.1}{5.2} & \fullres{-1.5}{2.4} & 48.1 \\

Auto-L & \fullres{41.6}{5.4} & \fullres{42.3}{6.2} & \fullres{84.3}{3.3} & \fullres{57.9}{3.9} & \fullres{\textbf{61.9}}{7.5} & \fullres{\textbf{67.7}}{7.9} & \fullres{55.5}{5.0} & \fullres{1.2}{4.8} & 51.6 \\
AMuLaP & \fullres{50.8}{2.1} & \fullres{52.3}{1.8} & \fullres{\textbf{86.9}}{1.6} & \fullres{53.1}{2.8} & \fullres{58.9}{7.9} & \fullres{56.3}{5.0} & \fullres{60.2}{2.7} & \fullres{2.3}{1.4} & 52.6  \\
Few-shot fine-tuning & \fullres{45.8}{6.4} & \fullres{47.8}{6.8} & \fullres{81.4}{3.8} & \fullres{\textbf{60.2}}{6.5} & \fullres{54.4}{3.9} & \fullres{76.6}{2.5} & \fullres{\textbf{60.7}}{4.3} & \fullres{\textbf{33.9}}{14.3}  & \textbf{57.6} \\
\hline
\multicolumn{10}{l}{\textit{Fully zero-shot}}\\
Majority & 32.7 & 33.0 & 50.9 & 49.5 & 52.7 & \textbf{81.2} & 0.0 & 0.0 & 37.5 \\
Null Prompt & \fullres{33.1}{0.4} & \fullres{33.8}{0.5} & \fullres{79.1}{4.0} & \fullres{50.7}{0.1} & \fullres{47.2}{0.6} & \fullres{12.9}{7.0} & \fullres{1.3}{1.0} & \fullres{-1.1}{2.0} & 32.1 \\
Multi-Null Prompt & \fullres{38.0}{3.5} & \fullres{38.5}{4.1} & \fullres{70.2}{7.7} & \fullres{52.2}{1.7} & \fullres{53.0}{2.2} & \fullres{19.9}{8.7} & \fullres{25.5}{13.4} & \fullres{\textbf{6.2}}{2.0} & 37.9 \\
\method  & \fullres{\textbf{45.7}}{0.6} & \fullres{\textbf{45.9}}{0.5} & \fullres{\textbf{86.3}}{1.2} & \fullres{\textbf{57.6}}{0.7} & \fullres{\textbf{55.0}}{3.4} & \fullres{79.8}{1.6} & \fullres{\textbf{52.4}}{0.4} & \fullres{4.9}{4.1} & \textbf{53.5} \\
\Xhline{2\arrayrulewidth} 
\end{tabular}
}
\caption{The performance of \method with RoBERTa-large on GLUE benchmark against other methods, including few-shot learning methods. Manual Label: using the human-designed prompts in \cite{Gao2021MakingPL}; In-context learning: using the in-context learning proposed in \cite{Brown2020LanguageMA} with RoBERTa-large; Auto-L: method in \cite{Gao2021MakingPL}; AMuLaP: method in \cite{Wang2022AutomaticMP}; Majority: majority class.}
\label{table:glue}
\end{table*}

\subsection{Baselines}
We evaluate the following baseline methods.
\paragraph{Semantic Retrieval} We utilize sentence embedding models \citep{reimers-2019-sentence-bert} to obtain the embedding for each sentence and descriptions for each class. Then we calculate the cosine similarity between sentences and label descriptions. We assign the most similar class labels to the sentence. Particularly, we use \texttt{all-mpnet-base-v2} from Hugging Face 
as the sentence embedding model, and the descriptions for each class can be found in Appendix \ref{sec:app_exp}.
\paragraph{NSP-BERT} \cite{Sun2021NSPBERTAP} propose text entailment tasks to replace text classification tasks and then use the Next Sentence Prediction (NSP) head to predict the results. We show the template we use in Appendix \ref{sec:app_exp}.
\paragraph{ManualVerb} Manual verbalizers are defined by human experts with domain knowledge and we simply use the label words provided by OpenPrompt \citep{ding2021openprompt}.
\paragraph{LOTClass} \cite{Meng2020TextCU} employ pre-trained neural language models with unlabeled data for category understanding, i.e., finding words similar to label names. They then introduce a self-training approach to the entire unlabeled corpus to generalize the model.
\paragraph{GPT-3 with descriptions} Following \cite{Brown2020LanguageMA}, we manually write the descriptions for each class and query GPT-3 where the predicted token serves as the prediction. We show the descriptions in Appendix \ref{sec:app_exp}.

\paragraph{ChatGPT with descriptions} In the case of ChatGPT  \cite{OpenAI2022ChatGPT}, we employ the same descriptions as those used for GPT-3. We query the ChatGPT model using these descriptions, and the predicted token is considered as the corresponding prediction. Our experimentation is based on the March 2023 version of ChatGPT.

\paragraph{SimPTC} \citet{Fei2022BeyondPM} show that zero-shot text classification can be improved by leveraging text clustering in the embedding spaces of pre-trained language models. SimPTC utilizes a Bayesian Gaussian Mixture Model to fit unlabeled texts. The initialization of cluster positions and shapes is performed using class names.

\paragraph{KPT} \cite{Hu2022KnowledgeablePI} propose knowledgeable prompt-tuning, which expands the label words space using external knowledge bases (KB). KPT also refines the expanded label words based on the unlabeled data. We show the best results of KPT in the zero-shot setting.
\paragraph{Null Prompt}\cite{LoganIV2022CuttingDO} insert a  token at the end of the text (i.e. using the prompt template `` \texttt{[$x$][MASK]}" ) and then use the prediction of the \texttt{[MASK]} token to perform zero-shot classification.
\paragraph{Multi-Null prompting} \cite{Wang2021Null} find that simply introducing a few prompt \texttt{[MASK]}s can improve the performance and robustness of the Null Prompt in the zero-shot settings.

\subsection{Main Results}
We demonstrate our experimental results in Table \ref{table:main_res}. Overall \method outperforms Null Prompt and Multi-Null Prompt remarkably by over 10 percent in a fully zero-shot setting. \method achieves an accuracy of over 85\% on AG News and DBPedia and over 90\% on IMDB and Amazon. We conjecture that topic classifications in AG News and DBPedia are more complicated than binary sentiment classifications in IMDB and Amazon, hence the higher accuracy on the latter. 

\method is only slightly worse than KPT and SimPTC but outperforms most baseline methods in which human efforts/external knowledge or unlabeled data are strictly required. It's worth noting that \method performs much better than ManualVerb, suggesting that the label words generated by our method are more comprehensive and unbiased than human-designed ones. Besides, \method can beat GPT-3 by 4\% in terms of average accuracy, a strong sign of the great potential for RoBERTa-large with 355M parameters compared to 175B parameters giant GPT-3.

To explore how our method \method performs on different kinds of tasks, we also conduct experiments on the GLUE benchmark \citep{Wang2018GLUEAM}. Specifically, we test on Multi-Genre Natural Language Inference Matched (MNLI), Multi-Genre Natural Language Inference Mismatched (MNLI-mm)\citep{Williams2018ABC} , Question Natural Language Inference (QNLI) \citep{Rajpurkar2016SQuAD1Q} and Recognizing Textual Entailment (RTE) \citep{Bentivogli2009TheSP} for Natural Language Inference (NLI); Microsoft Research Paraphrase Matching (MRPC) \citep{Dolan2005AutomaticallyCA} and Quora Question Pairs (QQP) \citep{chen2018} for Paraphrase Similarity Matching; Stanford Sentiment Treebank (SST-2) \citep{Socher2013RecursiveDM} for Sentiment Classification; The Corpus of Linguistic Acceptability (CoLA) \citep{Warstadt2019NeuralNA} for Linguistic Acceptability.

As shown in Table \ref{table:glue}, \method outperforms all other methods in fully zero-shot setting. Auto-L \citep{Gao2021MakingPL} and AMuLaP \citep{Wang2022AutomaticMP} are both automatic label words searching methods utilizing few-shot examples. Our method \method can even outperform them without any unlabeled data or few-shot training examples. 


\subsection{Effects of similarity functions in nonparametric aggregation}

\begin{figure}[htbp]
\centering
\includegraphics[width=0.43\textwidth]{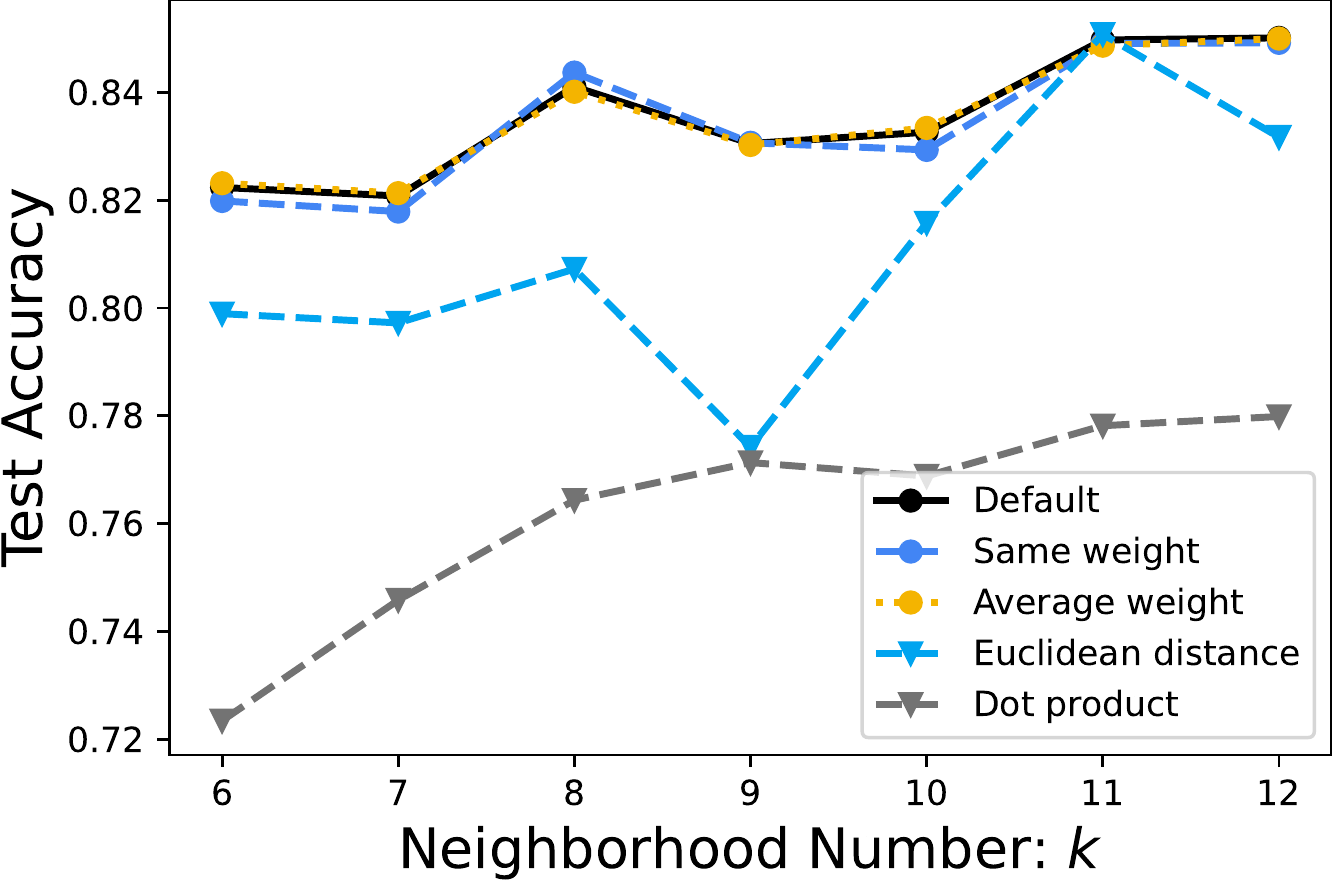}
\caption{Effects of different aggregation.}
\label{fig:res_weight}
\end{figure}
Both weight and similarity functions play a critical role in the design of \method and we test how \method performs on AG News with different configurations. The ``Default" setting is as stated in Equation \ref{eq:topk} and \ref{eq:weight}. We fix the similarity function $ S\left(\mathbf{emb}({v_i}), \mathbf{emb}({y_j})\right)=\frac{\mathbf{emb}({v_i})}{\left\|\mathbf{emb}({v_i})\right\|} \cdot \frac{\mathbf{emb}({y_j})}{\left\|\mathbf{emb}({y_j})\right\|}$, set $w(v_i, y_j) = 1$ for the ``Same weight" setting and $w(v_i, y_j) = \frac{S(\mathbf{emb}({v_i}), \mathbf{emb}({y_j}))}{\sum_{v_k \in \mathcal{M}(y_j)} S(\mathbf{emb}({v_k}), \mathbf{emb}({y_j}))}$ for the ``Average weight" setting. Besides cosine similarity, the Euclidean distance and the dot product are also common similarity measures for embeddings. Consequently, we fix the weight $w(v_i, y_j) = 1$, choose $S\left(\mathbf{emb}({v_i}), \mathbf{emb}({y_j})\right)=-\|\mathbf{emb}({v_i}) - \mathbf{emb}({y_j})\|$ for the ``Euclidean distance" setting and $S\left(\mathbf{emb}({v_i}), \mathbf{emb}({y_j})\right)=\mathbf{emb}({v_i})\cdot \mathbf{emb}({y_j})$ for the ``Dot product" setting. It can be informed from Figure \ref{fig:res_weight} that with a fixed similarity function, different weight calculations yield comparable results, but with a fixed weight, cosine similarity is the optimal similarity measure. 
\begin{figure}[h]
\centering
\includegraphics[width=0.43\textwidth]{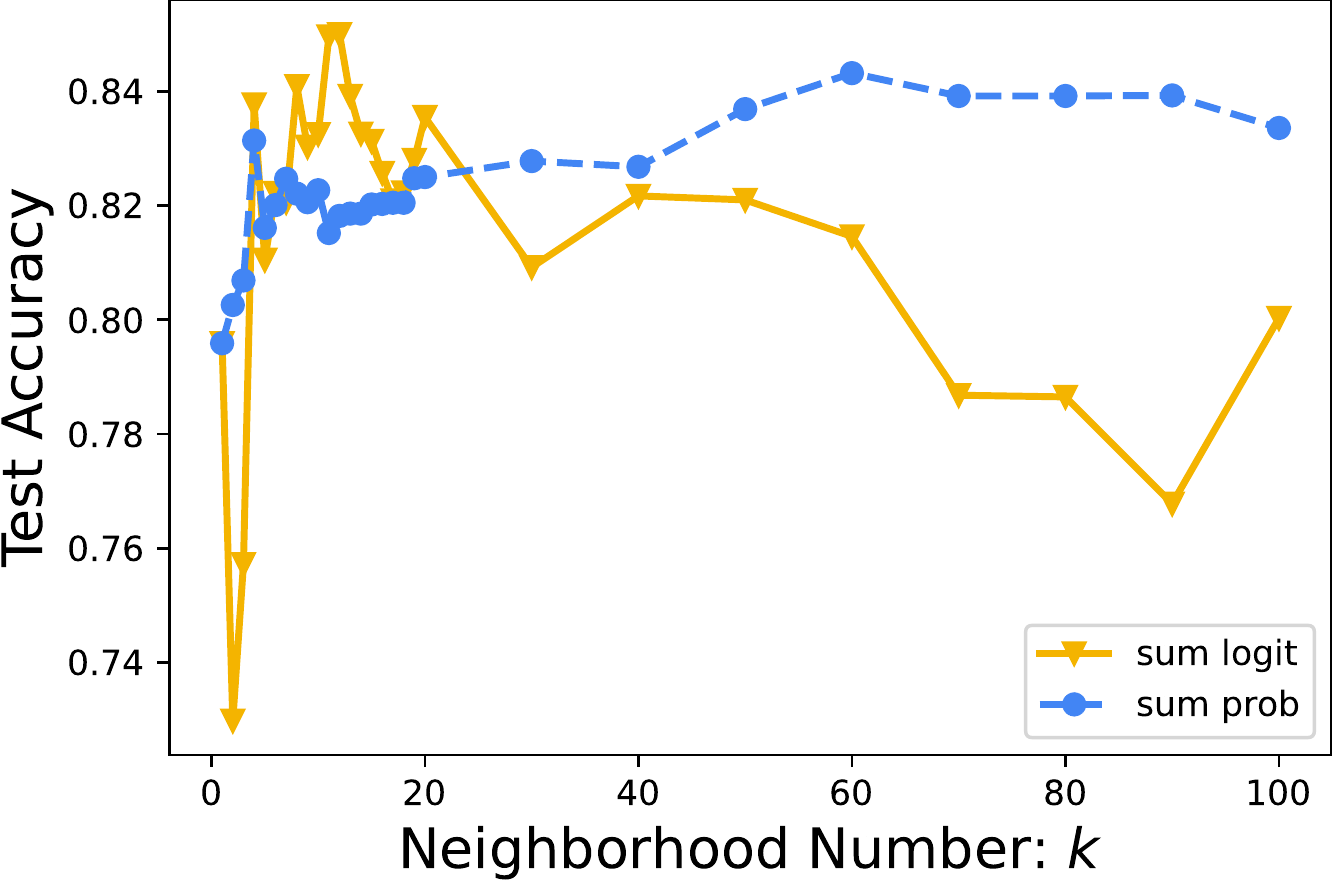}
\caption{Test results on AG News.}
\label{fig:prob_logit}
\end{figure}

\begin{figure*}[htbp]
\centering
\includegraphics[width=0.73\textwidth]{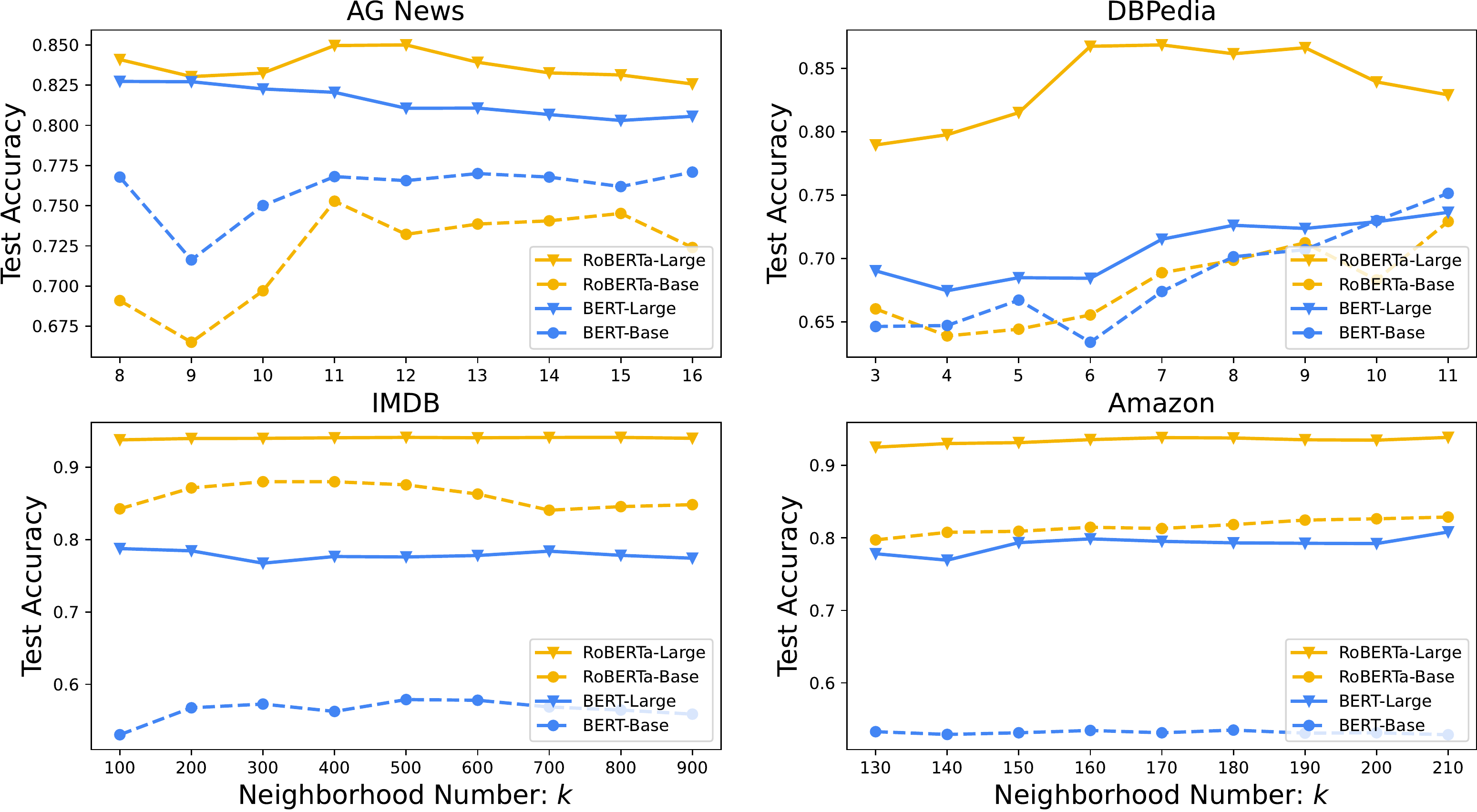}
\caption{Test results of \method for four PLMs with different neighborhood numbers.}
\label{fig:select}
\end{figure*}

\subsection{Can we sum over probabilities?}

\method sums up all logits for a label word set as shown in Equation \ref{eq:logits}. Another possible approach is to sum up the probabilities from PLM's prediction for the label words and choose the argmax for all different labels as the prediction:
$P(y_j|x_{\mathrm{prompt}}) = \sum_{v_i\in\mathcal{M}(y_j)}w(v_i, y_j) \cdot P(\texttt{[MASK]} = v_i| x_{\mathrm{prompt}})$,
$\widetilde{y}=\underset{y_j}{\arg \max}~P\left(y_j \mid x_{\mathrm{prompt}}\right)
$.
We conduct experiments on AG News to compare the above two approaches, one that sums up logits (``sum logit") and one that sums up probabilities (``sum prob"). Figure \ref{fig:prob_logit} presents the results and we find that ``sum logit" performs better at small $k$ but ``sum prob" delivers better results when $k$ exceeds 30. ``sum logit" achieves the best result at $k=12$ among all experiments.

\subsection{How many label words should we choose?}

The number of label words impacts the performance of our method \method as well. In Figure \ref{fig:select}, we display the performances of different models with varied neighborhood numbers. In general, \method attains similar test accuracy across different neighborhood numbers. Regardless of the choice for neighborhood number, \method-RoBERTa-large achieves over 80\% accuracy in topic classification tasks on AG News and DBPedia, and it gains over 90\% accuracy in sentiment classification tasks on IMDB and Amazon. In real-world applications, we can simply choose a fixed neighborhood number (e.g. 8-10) to achieve decent performance.

\hide{Typically, the variance of the performance is small in a fixed window size}

\subsection{How does \method perform with different PLMs?}

\begin{table}[htbp]
\setlength{\tabcolsep}{3pt}
\small
\centering
\begin{tabular}{lccccc}
\Xhline{2\arrayrulewidth} 
\textbf{Method} & \textbf{AG} & \textbf{DB}  & \textbf{IM}  &\textbf{AZ} & \textbf{Avg.}  \\
\hline
\method-T5-base & 76.8 & 78.3 & 68.5 & 65.3 & 72.2 \\
\method-GPT2-base & 81.1 & 78.1 & 83.7 & 85.6 & 82.1 \\
\method-BERT-base        & 79.4  & 77.8  & 57.7  & 53.5  & 67.1 \\
\method-BERT-large       & 82.7  & 80.9  & 81.6   & 80.8  & 81.5 \\    
\method-RoBERTa-base     & 75.3   & 82.8   & 88.7 & 83.9  & 82.7\\
\method-RoBERTa-large    & \textbf{85.0}  & \textbf{86.8}  & \textbf{94.1}  & \textbf{93.9} & \textbf{90.0} \\
\Xhline{2\arrayrulewidth} 
\end{tabular}
\caption{The zero-shot results of different backbones. \method-RoBERTa-large performs the best in all datasets. AG: AG News; DB: DBPeida; IM: IMDB; AZ: Amazon.}
\label{table:diff_model}
\end{table}

The performance of \method heavily relies on the choice of the pre-trained language model. This is due to the variations in label words for different categories, which stem from the distinct initial word embeddings and vocabularies employed by each PLM. Additionally, \method can be adapted for text generation models such as T5 \citep{JMLR:v21:20-074} and GPT-2 \citep{radford2019language}) with minor modifications. In our approach, we utilize T5-base/GPT2-base to generate the missing spans at the end of the prompt text. The first predicted token serves as the input to the verbalizer, and we follow the nonparametric aggregation steps outlined in Appendix \ref{sec:app_exp} to determine the category. 

To investigate the impact of employing different PLMs, we conduct additional experiments using BERT-base-cased, BERT-large-cased, RoBERTa-base, T5-base, and GPT2-base models. The results are presented in Table \ref{table:diff_model}. Notably, \method with RoBERTa-large achieves the highest performance, which can be attributed to the model's extensive parameter count and the fact that it is pre-trained on a large corpus. As anticipated, larger models such as RoBERTa-large and BERT-large outperform their base counterparts (RoBERTa-base and BERT-base) on average, with RoBERTa consistently exhibiting superior accuracy compared to BERT models. While \method-T5-base and \method-GPT2-base demonstrate commendable performance, they do not surpass the performance of \method-RoBERTa-large.

\subsection{Is \method limited to text classification tasks}
\begin{table}[h]
\small
\centering
\begin{tabular}{lc} \\ 
\Xhline{2\arrayrulewidth}   
\textbf{Method} & \textbf{CQA Dev Set Accuracy} \\ 
\hline
Few-shot Direct GPT-J  & 20.9 \\
Few-shot CoT GPT-J & 36.6 \\
Few-shot CoT LaMDA 137B & 55.6 \\
\hline
NPPrompt-RoBERTa-large & 34.2 \\
\Xhline{2\arrayrulewidth} 
\end{tabular}
\rebuttal{\caption{Test results on CommonsenseQA dataset. Direct: directly output the final answer; CoT: prompted with chain-of-thought (CoT) rationales; LaMDA: method in \cite{Wei2022ChainOT}.}\label{table:qa}}
\vspace{-5mm}
\end{table}
Our research extends beyond text classification and encompasses experiments on multiple-choice question answering (QA) tasks as well. Specifically, we assess the performance of \method using the widely-utilized CommonsenseQA (CQA) dataset \citep{Talmor2019CommonsenseQAAQ}. In this new setting, we use the prompt template ``$x$ The answer is \texttt{[MASK]}.”, e.g. ``What do animals do when an enemy is approaching? The answer is \texttt{[MASK]}.”. Subsequently, we search for the $k$-nearest neighbors for each target answer, setting $k$ as 15. The prediction is obtained by applying the same process employed for text classification tasks. The results of our experiments are presented in Table \ref{table:qa} (few-shot results obtained from \cite{Zelikman2022STaRBR}). Notably, NPPrompt not only achieves satisfactory performance on the CommonsenseQA dataset but even outperforms few-shot GPT-J \citep{mesh-transformer-jax} as well. This demonstrates the versatility and flexibility of \method across various NLP scenarios.

\section{Discussion}
\label{sec:dis}


Our proposed method, \method, demonstrates exceptional performance in zero-shot text classification tasks. We attribute this success to two key factors. Firstly, by utilizing the initial word embedding from pre-trained language models (PLMs), we are able to identify cognates of the label words. For instance, in Table \ref{table:sim}, we observe variations of the word "business" such as "Business" and "businesses" for the \textsc{Business} category. Secondly, we effectively leverage the capabilities of pre-trained language models by reformulating the zero-shot classification problem as a masked token prediction task, which aligns with the pre-training process.

Furthermore, \method offers a promising solution for dynamic and open zero-shot classification problems, where new classes may arise or old classes may be removed. With the use of efficient PLMs and category names, as well as the key word design in Equation \ref{eq:key_words}, \method can also be applied in scenarios where label names do not possess semantic meaning (e.g. categories with label names {``A'', ``B'', ``C''}). This technique has the potential for wide deployment in real-world applications.

\section{Conclusion}
\label{sec:conclusion}
In this paper, we propose \method, a novel and effective method for fully zero-shot learning with pre-trained language models. We use initial word embedding of PLM to automatically find related words for category names, which enables us to construct the verbalizers without manual design or unlabeled corpus. Experimental results show that \method outperforms the previous zero-shot methods by large margins.

\section*{Limitations}
For those label names without semantic meanings, several keywords are still required for \method to work well. Furthermore, this study focuses exclusively on the zero-shot setting. However, there are potential avenues for exploration in the few-shot scenario, which is prevalent in practical applications. The applicability of \method to other tasks, such as ranking and relation extraction, remains uncertain and warrants further investigation. Designing a refinement method to jointly search for label words and templates can be a promising direction for future research.


\bibliography{anthology, custom}
\bibliographystyle{acl_natbib}

\appendix

\newpage


\section{Appendix}
\label{sec:appendix}

\subsection{Experimental Details}
\label{sec:app_exp}

Table \ref{tab:nsp_bert} shows all the manual templates of NSP-BERT. We show the prompt templates for \method-T5 in Table \ref{tab:t5}. Table \ref{tab:semantic} summarizes manual designed descriptions of each dataset for Semantic Retrieval. As for GPT-3, we query the OpenAI API\footnote{\url{https://openai.com/api/}} and test with \texttt{Davinci-001} model. The prompts for GPT-3 are shown in Table \ref{tab:gpt3}. We list all templates and label names for \method of all experiments in Table \ref{tab:glue_temp}.
We also list the related words result in sentiment classification (\textsc{Good}/\textsc{Bad}) and NLI (\textsc{Yes}/\textsc{No})) tasks in Table \ref{table:sim_new}.

\begin{table}[h!]
\vspace{-4mm}
\setlength{\tabcolsep}{2pt} 
\centering
\begin{tabular}{cc} \\ 
\Xhline{2\arrayrulewidth}   
\textbf{Dataset} & \textbf{Template}  \\ 
\hline
AG News & News: \textit{label name}. \\
DBPedia & News: \textit{label name}. \\
IMDB & This text shows \textit{label name} sentiment.\\
Amazon & The attitude of this text is \textit{label name}.  \\
\Xhline{2\arrayrulewidth} 
\end{tabular}
\caption{Prompt templates of NSP-BERT \citep{Sun2021NSPBERTAP} in Table \ref{table:main_res}.}
\label{tab:nsp_bert}
\vspace{-6mm}
\end{table}

\begin{table}[h]
\small
\centering
\begin{tabular}{llc} \\ 
\Xhline{2\arrayrulewidth}   
\textbf{Dataset} & \textbf{Template}\\ 
\hline
AG News & $x$ In this sentence, the topic is about \mask  \\
DBPedia & $x_1$ $x_2$ In this sentence, $x_1$ is a \mask  \\
IMDB & $x$ In summary, the movie was \mask \\
Amazon & $x$ All in all, it was \mask  \\
\Xhline{2\arrayrulewidth} 
\end{tabular}
\caption{Prompt template of \method with T5-base ($k=15$) in Tabel \ref{table:diff_model}.}\label{tab:t5}
\vspace{-5mm}
\end{table}


\subsection{What label words do different PLMs choose?}
\begin{table}[h]
\setlength{\tabcolsep}{2pt}
\tiny
\centering
\begin{tabular}{llllllll}
\Xhline{2\arrayrulewidth} 
\multicolumn{2}{c}{\textbf{RoBERTa-large}} & \multicolumn{2}{c}{\textbf{RoBERTa-base}} & \multicolumn{2}{c}{\textbf{BERT-large}} & \multicolumn{2}{c}{\textbf{BERT-base}} \\
\multicolumn{1}{c}{Word} & \multicolumn{1}{c}{Sim} & \multicolumn{1}{c}{Word} & \multicolumn{1}{c}{Sim}  & \multicolumn{1}{c}{Word} & \multicolumn{1}{c}{Sim}  & \multicolumn{1}{c}{Word} & \multicolumn{1}{c}{Sim}  \\
\hline
`` school" & 1.00 & `` school" & 1.00 & ``school" & 1.00 & ``school" & 1.00 \\
`` School" & 0.80  & `` School" & 0.75 & ``School" & 0.69 & ``School" & 0.70 \\
`` schools" & 0.77 & `` schools" & 0.71 & ``schools" & 0.63 & ``schools" & 0.63 \\
``school" & 0.74 & ``school" & 0.70 & ``college" & 0.55 & ``college" & 0.54 \\
`` SCHOOL" & 0.69 & ``School" & 0.70 & ``university" & 0.50 & ``university" & 0.51 \\
``School" & 0.68 &  `` SCHOOL" & 0.56 & ``student" & 0.42 & ``College" & 0.40 \\
`` university" & 0.66 & `` college" & 0.50 & ``church" & 0.41 & ``church" & 0.40 \\
`` college" & 0.65 &  `` university" & 0.50 & ``house" & 0.38 & ``student" & 0.37 \\
`` Schools" & 0.65 & `` Schools" & 0.49 & ``education" & 0.38 & ``students" & 0.37 \\
`` schooling" & 0.64 & `` schooling" & 0.45 & ``students" & 0.37 & ``Schools" & 0.37 \\
`` preschool" & 0.63  & `` preschool" & 0.44 & ``class" & 0.37 & ``academy" & 0.37 \\
`` kindergarten" & 0.63 &  `` kindergarten" & 0.41& ``town" & 0.37 & ``class" & 0.36 \\
`` classroom" & 0.60  & `` student" & 0.41 & ``College" & 0.36 & ``education" & 0.36 \\
`` student" & 0.58 & `` students" & 0.39 & ``Schools" & 0.36 & ``University" & 0.35 \\
`` education" & 0.58 & `` classroom" & 0.38 & ``work" & 0.35 & ``house" & 0.35 \\
\Xhline{2\arrayrulewidth} 
\end{tabular}
\caption{The top 15 similar words of \textsc{School} category in the DBPedia dataset. Sim: similarity scores.}
\vspace{-4mm}
\label{table:sim_diff}
\end{table}
We summarize the label words of different PLMs for \textsc{School} category in DBPedia in Table \ref{table:sim_diff}. RoBERTa-large and RoBERTa-base share similar sets of label words yet with a minor discrepancy between their similarity scores. RoBERTa-large usually produces larger similarities than RoBERTa-base. In contrast, the label words in RoBERTa are quite different from those in BERT.

\rebuttal{\subsection{Extension to Multi-Word Expressions}}
\label{sec:multi-word}
Here we extend our method to support multi-word label names like \textsc{NaturalPlace}, \textsc{MeanOfTransportation} and etc. The major part is to obtain related words to a multi-word label name. Once we obtain the related words, the rest non-parametric aggregation step remains identical. 
We consider two scenarios:

\paragraph{The label name is multi-word (i.e., phrase) and related words are still single-words} To model the phrase, we use average contextualized embedding instead of word embedding for both label names and related single-words to compute cosine similarity. As suggested in \cite{whiten}, we whiten the contextualized output of RoBERTa by a linear transformation obtained from the contextualized embedding of all words in vocabulary. To obtain the best result, we select the output of layer 6 of RoBERTa. This extension achieves 61\% accuracy on the DBPedia dataset using the original multi-word label names (original label names can be found at \url{https://rdrr.io/cran/textdata/man/dataset_dbpedia.html}).

\paragraph{Both the label name and related words are phrases} Since the search space of a related phrase is exponentially large in its length, we use another prompt to filter candidate words. The template we use is ``\texttt{[LABEL\_NAME]} can also be called \texttt{[MASK]}$*n$.'', where $n$ is the length of the candidate. For example, if the label name is \textsc{MeanOfTransportation} and $n=2$, the template will look like ``Mean of transportation can also be called \texttt{[MASK]} \texttt{[MASK]}.''. We feed it to RoBERTa and filter top-$k$ candidate phrases of masked prediction. Since masked prediction is conditionally independent of each mask, we further re-rank the top-$k$ candidate phrases by either the contextualized embedding method mentioned above or another auto-regressive LM. For the latter one, we evaluate the perplexity of the template with \texttt{[MASK]} filled by candidate phrases. This generates 71\% accuracy on DBPedia if the length of the phrase is two and the re-ranking is performed by GPT-2 \citep{radford2019language}. 

\begin{table*}[h!]
\small
\centering
\begin{tabular}{l} \\ 
\Xhline{2\arrayrulewidth}   
\textbf{Descriptions}  \\ 
\hline
\textit{AG News}: \\
The politics category is related to politics, government, and law. \\
The sports category is related to sports, competition, and athletics. \\
The business category is related to business, portfolio, economics, and money. \\
The technology category is related to technology, software, system, and science. \\
\hline
\textit{DBPedia}: \\
The company category is related to company, corporation, enterprise, brand, and business.\\
The school category is related to school, academy, university, and college.\\
The artist category is related to artist, art, painter, musician, singer, and creative.\\
The athlete category is related to athletes, sports, Olympic, and gym.\\
The politics category is related to politics, government, and law.\\
The transportation category is related to transportation, transport, vehicle, and traffic.\\
The building category is related to buildings, construction, and structure.\\
The mountain category is related to river, lake, bay, and mountain.\\
The village category is related to village, town, and rural.\\
The animal category is related to animal, wildlife, and nature.\\
The plant category is related to plant, shrub, tree, and forest.\\
The album category is related to album, lyrics, cd, and song.\\
The film category is related to film, movie, cinema, and video.\\
The book category is related to book, novel, and publication.\\
\hline
\textit{IMDB}:\\
The bad category is related to negative and bad reviews.\\
The good category is related to positive and good reviews.\\
\hline
\textit{Amazon}:\\
The bad category is related to negative and bad reviews.\\
The good category is related to positive and good reviews.\\
\Xhline{2\arrayrulewidth} 
\end{tabular}
\caption{Descriptions for Semantic Retrieval in Table \ref{table:main_res}.}
\label{tab:semantic}
\end{table*}

\begin{table*}[htbp]
\small
\centering
\begin{tabularx}{\textwidth}{X} \\ 
\Xhline{2\arrayrulewidth}   
\textbf{Prompts for GPT-3 and ChatGPT}  \\ 
\hline
\textit{AG News} : \\
\texttt{[Descriptions]} Definition: In this task, you are given a sentence. Your job is to classify the following sentence into one of the four different categories. The categories are: ``politics'', ``sports'', ``business'', and ``technology''. Input: \texttt{[x]}. Output: \\ \hline
\textit{DBPedia}: \\
\texttt{[Descriptions]} Definition: In this task, you are given a sentence. Your job is to classify the following sentence into one of the fourteen different categories. The categories are: ``company'', ``school'', ``artist'', ``athlete'', ``politics'', ``transportation'', ``building'', ``mountain'', ``village'', ``animal'', ``plant'', ``album'', ``film'', and ``book''. Input: \texttt{[x]}. Output: \\ \hline
\textit{IMDB}:\\
\texttt{[Descriptions]} Definition: In this task, you are given a sentence. Your job is to classify the following sentence into one of the two categories. The categories are: ``bad" and ``good". Input: \texttt{[x]}. Output: \\ \hline
\textit{Amazon}:\\
\texttt{[Descriptions]} Definition: In this task, you are given a sentence. Your job is to classify the following sentence into one of the two categories. The categories are: ``bad" and ``good". Input: \texttt{[x]}. Output: \\ \hline
\Xhline{2\arrayrulewidth} 
\end{tabularx}
\caption{Prompts for GPT-3 and ChatGPT with descriptions \texttt{[Descriptions]} from Table \ref{tab:semantic} and input text \texttt{[x]}.}
\label{tab:gpt3}
\vspace{15mm}
\end{table*}

\begin{table*}[htbp]
\centering
\begin{tabular}{lllc} \\ 
\Xhline{2\arrayrulewidth}   
\textbf{Dataset} & \textbf{Template} & \textbf{Label Names} & \textbf{$k$}  \\ 
\hline
\multirow{4}*{AG News} & \multirow{4}*{A \texttt{[MASK]} news : $x$ .} & category 1: \textit{world, politics} & \multirow{4}*{12} \\
&&category 2: \textit{sports} & \\
&&category 3: \textit{business} & \\
&&category 4: \textit{technology, science} & \\ \hline
\multirow{14}*{DBPedia} & &category 1: \textit{company} &\multirow{14}*{7} \\ 
&&category 2: \textit{school}& \\
&&category 3: \textit{artist}& \\
&&category 4: \textit{sports}& \\
&&category 5: \textit{politics, office}& \\
&&category 6: \textit{transportation, car, bus, train}& \\
&$x_1$ $x_2$ In this sentence, $x_1$ &category 7: \textit{building, construct, room, tower}& \\
&is a \texttt{[MASK]} . &category 8: \textit{river, lake, mountain}& \\
&&category 9: \textit{village}& \\
&&category 10: \textit{animal, pet}& \\
&&category 11: \textit{plant}& \\
&&category 12: \textit{album}& \\
&&category 13: \textit{film}& \\
&&category 14: \textit{book, publication}& \\ \hline
\multirow{2}*{IMDB} & \multirow{2}*{$x$ All in all, it was \texttt{[MASK]} .} & positive: \textit{good} & \multirow{2}*{500} \\
&&negative: \textit{bad}& \\ \hline
\multirow{2}*{Amazon} & \multirow{2}*{$x$ All in all, it was \texttt{[MASK]} .}  & positive: \textit{good} & \multirow{2}*{170} \\
&&negative: \textit{bad}& \\ \hline
\multirow{2}*{SST-2} & \multirow{2}*{$x_1$ It was \texttt{[MASK]} .}    & positive: \textit{great} & \multirow{2}*{9} \\
&&negative: \textit{terrible}& \\ \hline
\multirow{3}*{MNLI}       & \multirow{3}*{$x_1$ ? \texttt{[MASK]} , $x_2$}   & entailment: \textit{yes} & \multirow{3}*{4} \\
&&neutral: \textit{maybe}&\\
&&contradiction: \textit{no}&\\ \hline
\multirow{3}*{MNLI-mm}    & \multirow{3}*{$x_1$ ? \texttt{[MASK]} , $x_2$}   & entailment: \textit{yes} & \multirow{3}*{4} \\
&&neutral: \textit{maybe}&\\
&&contradiction: \textit{no}&\\ \hline
\multirow{2}*{QNLI}       & \multirow{2}*{$x_1$ ? \texttt{[MASK]} , $x_2$}   & entailment: \textit{Yes, Indeed, Overall}     & \multirow{2}*{3} \\
&&not\_entailment: \textit{No, Well, However}&\\ \hline 
\multirow{2}*{RTE}        & \multirow{2}*{$x_1$ ? \texttt{[MASK]} , $x_2$}   & entailment: \textit{Yes} & \multirow{2}*{10}\\
&& not\_entailment: \textit{No} & \\ \hline
\multirow{2}*{MRPC}       & \multirow{2}*{$x_1$ \texttt{[MASK]} , $x_2$}     & equivalent: \textit{Yes} & \multirow{2}*{9} \\
&& not\_equivalent: \textit{No} & \\ \hline
\multirow{2}*{QQP}        & \multirow{2}*{$x_1$ \texttt{[MASK]} , $x_2$}     & equivalent: \textit{Yes} & \multirow{2}*{9} \\
&&not\_equivalent: \textit{No}& \\ \hline
\multirow{2}*{CoLA}       & \multirow{2}*{$x_1$ This is \texttt{[MASK]} .}   & grammatical: \textit{true}& \multirow{2}*{7} \\
&&not\_grammatical: \textit{wrong}& \\                          
\Xhline{2\arrayrulewidth} 
\end{tabular}
\caption{Templates and label names for \method. $k$ refers to the best neighborhood number for RoBERTa-large.}
\label{tab:glue_temp}
\end{table*}

\begin{table*}[h]
\small
\centering
\begin{tabular}{llllllll}
\Xhline{2\arrayrulewidth} 
\multicolumn{2}{c}{\textsc{Good}} & \multicolumn{2}{c}{\textsc{Bad}} & \multicolumn{2}{c}{\textsc{Yes}} & \multicolumn{2}{c}{\textsc{No}} \\
\multicolumn{1}{c}{Word} & \multicolumn{1}{c}{Sim} & \multicolumn{1}{c}{Word} & \multicolumn{1}{c}{Sim}  & \multicolumn{1}{c}{Word} & \multicolumn{1}{c}{Sim}  & \multicolumn{1}{c}{Word} & \multicolumn{1}{c}{Sim}  \\
\hline
`` good" & 1.00 & `` bad" & 1.00 & `` Yes" & 1.00 & `` No" & 1.00 \\ 
`` Good" & 0.73 & `` Bad" & 0.71 & `` yes" & 0.79 & `` no" & 0.80 \\ 
`` GOOD" & 0.72 & `` terrible" & 0.69 & `` YES" & 0.73 & ``No" & 0.74 \\ 
``good" & 0.69 & `` BAD" & 0.69 & ``Yes" & 0.72 & `` NO" & 0.70 \\ 
`` great" & 0.66 & `` horrible" & 0.68 & `` Yeah" & 0.72 & `` Nope" & 0.62 \\ 
`` excellent" & 0.66 & ``bad" & 0.65 & `` Yep" & 0.65 & `` Yes" & 0.62 \\ 
`` decent" & 0.66 & `` awful" & 0.64 & `` Sure" & 0.62 & ``no" & 0.61 \\ 
``Good" & 0.65 & ``Bad" & 0.64 & `` No" & 0.62 & `` Nobody" & 0.59 \\ 
`` nice" & 0.64 & `` good" & 0.63 & `` Indeed" & 0.61 & `` Nos" & 0.57 \\ 
`` bad" & 0.63 & `` badly" & 0.62 & `` yeah" & 0.60 & `` The" & 0.57 \\ 
`` better" & 0.62 & `` crappy" & 0.60 & ``yes" & 0.59 & `` Yeah" & 0.57 \\ 
`` wonderful" & 0.58 & `` lousy" & 0.60 & `` Wow" & 0.59 & `` Nothing" & 0.56 \\ 
`` best" & 0.58 & `` worst" & 0.60 & `` Absolutely" & 0.58 & `` Not" & 0.56 \\ 
`` terrific" & 0.57 & `` horrendous" & 0.60 & `` Nope" & 0.58 & `` Never" & 0.56 \\ 
`` fantastic" & 0.57 & `` worse" & 0.59 & `` Okay" & 0.57 & `` None" & 0.55 \\ 
`` mediocre" & 0.57 & `` nasty" & 0.59 & `` Oh" & 0.57 & `` Number" & 0.55 \\ 
`` lousy" & 0.57 & `` shitty" & 0.59 & `` Hello" & 0.57 & `` So" & 0.54 \\ 
`` satisfactory" & 0.56 & `` dreadful" & 0.59 & `` Hey" & 0.57 & `` Any" & 0.54 \\ 
`` marvelous" & 0.56 & `` rotten" & 0.58 & `` Nevertheless" & 0.57 & `` And" & 0.54 \\ 
`` GREAT" & 0.56 & `` harmful" & 0.58 & `` However" & 0.56 & ``NO" & 0.53 \\ 
\Xhline{2\arrayrulewidth} 
\end{tabular}
\rebuttal{\caption{The top 20 similar words of label names in sentiment classification (\textsc{Good}/\textsc{Bad}) and NLI (\textsc{Yes}/\textsc{No}) tasks.}\label{table:sim_new}}
\end{table*}

\end{document}